\documentclass{article} % For LaTeX2e
\usepackage{iclr2027_conference,times}

\usepackage{amsmath,amsfonts,bm}

\def\eqref#1{equation~\ref{#1}}
\def\1{\bm{1}}

\DeclareMathAlphabet{\mathsfit}{\encodingdefault}{\sfdefault}{m}{sl}
\SetMathAlphabet{\mathsfit}{bold}{\encodingdefault}{\sfdefault}{bx}{n}

\usepackage{hyperref}
\usepackage{url}
\usepackage{graphicx}
\usepackage{booktabs}
\usepackage{multirow}
\usepackage{amsmath,amssymb}
\usepackage{xcolor}
\usepackage{colortbl}
\usepackage{float}
\usepackage{wrapfig}
\definecolor{tacdynblue}{HTML}{2F80ED}
\definecolor{tacdynviolet}{HTML}{9B51E0}
\title{%
\centering
\vspace*{-0.15in}% lift the arXiv title block while keeping it below the running head
{\normalfont\large\scshape
{\bfseries\textcolor{tacdynblue}{TacDyn}\textcolor{tacdynviolet}{-WAM}:}
Learning Implicit Tactile Dynamics in a\\[-1pt]
Heterogeneous Visuo-Tactile World Action Model}}

\author{%
{\fontsize{8}{9.5}\selectfont\bfseries
\makebox[\textwidth][c]{Enyi Wang$^{1}$\hspace{0.3em} Mingxin Wang$^{1,4}$\hspace{0.3em} Quan Shi$^{1}$\hspace{0.3em} Hetian Guo$^{1}$\hspace{0.3em}
Hongyu Wang$^{1}$\hspace{0.3em} Xi Wang$^{1}$\hspace{0.3em} Bin Qian$^{1,4}$}}\\[-1pt]
{\fontsize{8}{9.5}\selectfont\bfseries
\makebox[\textwidth][c]{Yupeng Zheng$^{2}$\hspace{0.3em} Wenxuan Song$^{3}$\hspace{0.3em} Houde Liu$^{4}$\hspace{0.3em} Yong Xu$^{1}$\hspace{0.3em}
Cheng Chi$^{5}$\hspace{0.3em} Wenchao Ding$^{6,7}$\hspace{0.3em} Yilun Chen$^{7}$\hspace{0.3em} Yan Wang$^{1\dagger}$}}\\[3pt]
{\normalfont\fontsize{8}{9.5}\selectfont
\makebox[\textwidth][c]{$^{1}$Institute for AI Industry Research (AIR), Tsinghua University \quad
$^{2}$Institute of Automation, Chinese Academy of Sciences}}\\[-1pt]
{\normalfont\fontsize{8}{9.5}\selectfont
\makebox[\textwidth][c]{$^{3}$The Hong Kong University of Science and Technology (Guangzhou) \quad
$^{4}$Tsinghua University}}\\[-1pt]
{\normalfont\fontsize{8}{9.5}\selectfont
\makebox[\textwidth][c]{$^{5}$School of Information, Renmin University of China \quad
$^{6}$Fudan University \quad $^{7}$TARS Robotics}}\\[-1pt]
{\normalfont\fontsize{8}{9.5}\selectfont
\makebox[\textwidth][c]{$^{\dagger}$Corresponding author}}
}

\hypersetup{
  pdftitle={TacDyn-WAM: Learning Implicit Tactile Dynamics in a Heterogeneous Visuo-Tactile World Action Model},
  pdfauthor={Enyi Wang, Mingxin Wang, Quan Shi, Hetian Guo, Hongyu Wang, Xi Wang, Bin Qian, Yupeng Zheng, Wenxuan Song, Houde Liu, Yong Xu, Cheng Chi, Wenchao Ding, Yilun Chen, Yan Wang}
}

\iclrfinalcopy % arXiv version: show authors
\begin{document}

\begingroup
\setlength{\tabcolsep}{0pt}
\maketitle
\endgroup
\lhead{Preprint}

\begin{abstract}
World action models improve robotic manipulation by conditioning actions on predicted futures, yet existing tactile variants largely inherit video-generation pipelines that reconstruct future tactile observations through iterative denoising.
Such prediction can become unreliable under deployment drift: small changes in contact position or force may substantially alter tactile pixels even when the underlying contact evolution remains predictable.
We introduce TacDyn-WAM, a heterogeneous visuo-tactile world action model that predicts implicit tactile dynamics rather than reconstructing future tactile observations.
It learns TacRep, a dynamics-aware tactile target space trained through masked spatio-temporal prediction on tactile clips and regularized by relational structure distillation.
A visual expert and an Implicit Tactile Dynamics Expert predict future visual and tactile representations in separate target spaces while interacting through joint attention; the tactile expert predicts future representations and their changes at multiple horizons in a single forward pass, and a read-only tactile memory supplies the current tactile state.
On UniVTAC, TacDyn-WAM achieves an average success rate of 81.5\% using only the provided demonstrations, reaching state-of-the-art-level performance and remaining competitive with models pretrained on large-scale visuo-tactile trajectories.
Ablations confirm the benefits of both tactile pathways and TacRep over pixel-reconstruction and static alternatives.
On five real-robot tasks, TacDyn-WAM reaches 71.0\% average success, and modest-scale pretraining raises it to 85.0\%, further validating our method.
{\hypersetup{hidelinks}\href{https://enyi-bean.github.io/TacDyn-WAM-Page/}{\textcolor{tacdynblue}{Project page}}}.
\end{abstract}

\section{Introduction}
\label{sec:intro}

World models forecast how observations evolve, while world action models use this foresight for policy learning, allowing actions to depend on predicted futures rather than current observations alone~\citep{cai2026internvla,li2026causal,yuan2026fast,ye2026gigaworld}.
Yet in existing world action models, this foresight comes primarily from vision.
In contact-rich or precision manipulation tasks, such as insertion and alignment under occlusion, cameras cannot fully resolve contact states, local geometry, or interaction forces.
Early efforts incorporated tactile sensing into VLA models, giving rise to vision-tactile-language-action (VTLA) models that encode tactile observations alongside vision and language~\citep{huang2025tactilevla,zhang2025vtla,bi2025vlatouch,cheng2025omnivtla}.
By understanding the current tactile state, they improve contact-rich manipulation over vision-only policies.
However, they do not predict how this state will evolve, leaving policies reactive to tactile feedback.
Recent tactile world action models close this gap by predicting future tactile observations as part of the world model~\citep{neoteai2026n0twam,lou2026dreamtac,yuan2026vtam,wu2026tactilewam}.
These models have achieved promising results, with some further benefiting from large-scale visuo-tactile pretraining.
Despite this progress, a core design question remains: what should tactile world models predict?

Most existing methods simply reuse video-generation pipelines: the same VAE encodes camera and tactile frames, while video backbones iteratively denoise future tactile latents.
This slows inference and ties prediction to pixel reconstruction, which is poorly suited to tactile observations.
Expert demonstrations cover only tactile images near correct contacts, whereas closed-loop errors may cause policies to drift from them.
This drift may leave camera views nearly unchanged, yet even slight changes in contact position or force can substantially alter tactile images, pushing them farther from the demonstration distribution.
Despite these pixel-level changes, contact evolution---whether an imprint deepens, slides, or rotates---often remains predictable.
We therefore argue for \emph{implicit tactile dynamics modeling}: predicting contact evolution in a dynamics-aware representation space rather than reconstructing future tactile observations.

Latent prediction is a step toward this goal, but not every latent space is equally suited to forecasting tactile evolution.
$N_0$-VTLA~\citep{neoteai2026n0vtla} predicts the net tactile change at a single horizon in the feature space of a frozen DINOv2~\citep{oquab2023dinov2}.
However, DINOv2 is trained on static natural images and is not explicitly optimized to capture frame-to-frame tactile evolution; recent tactile representation learning also shows the limitations of single-frame encoders in modeling such dynamics~\citep{feng2026anytouch}.
In contrast, V-JEPA~2.1~\citep{mur2026v} learns representations by predicting masked spatio-temporal features in videos, encouraging its encoder to capture predictable motion rather than exact pixel details.
Building on this objective, we learn a target space from tactile video clips that captures temporal change, while preserving spatial structure to support robust prediction under tactile distribution shifts.
Moreover, because a single net change can hide intermediate contact events when a squeeze or slip returns to a similar final state, we predict both future representations and their changes from the present at multiple horizons.
This captures contact evolution rather than only its endpoint.

In this paper, we present TacDyn-WAM, a heterogeneous visuo-tactile world action model whose two world-model experts use separate parameters and target spaces.
The visual expert predicts future visual latents, while the Implicit Tactile Dynamics Expert (ITDE) predicts tactile evolution in the TacRep space.
Joint attention couples their predictions at every layer and allows the Action Expert to use both.
The ITDE requires only a single forward pass without iterative denoising.
A separate Tactile Understanding Memory provides direct access to the current tactile state, while staged training progressively integrates the tactile modules with the pretrained experts.

On UniVTAC~\citep{chen2026univtac}, TacDyn-WAM achieves state-of-the-art-level performance using only the provided demonstrations.
It is competitive with the two strongest models on this benchmark, $N_0$-VTLA and $N_0$-TWAM, both of which rely on large-scale visuo-tactile pretraining.
Notably, $N_0$-TWAM follows the video-generation recipe and has roughly twice the parameters of TacDyn-WAM.
Ablations show the benefits of both tactile pathways and TacRep's advantage over pixel-reconstruction and static alternatives.
On five real-robot tasks, TacDyn-WAM trained on per-task demonstrations alone reaches 71.0\% average success, 16 points above the tactile policy FTP-1, and modest-scale pretraining on about 6,000 real visuo-tactile trajectories raises it to 85.0\%.

In summary, this work makes the following contributions:
\begin{itemize}
\item We propose TacDyn-WAM, a heterogeneous visuo-tactile world action model that jointly predicts future visual latents and tactile evolution in distinct target spaces, while a read-only Tactile Understanding Memory provides the current tactile state.
\item We introduce TacRep, a dynamics-aware representation space for implicit tactile dynamics prediction, enabling one-pass multi-horizon prediction without pixel reconstruction or iterative denoising.
\item TacDyn-WAM achieves state-of-the-art-level performance on UniVTAC using only the provided demonstrations, and ablations validate the design.
On a real robot, it outperforms vision-only and tactile baselines and benefits further from modest-scale tactile pretraining.
\end{itemize}

\section{Method}
\label{sec:method}

\subsection{Problem Setup and Overview}
\label{sec:overview}

\textbf{Base Model.}
We build on the 3B checkpoint of InternVLA-A1~\citep{cai2026internvla}, a mixture-of-transformers world action model with three experts coupled through block-wise causal joint attention.
The \emph{Understanding Expert}, a Qwen3-VL-2B vision-language model~\citep{bai2025qwen3}, encodes the current RGB observations $I_t$ and language instruction $\ell$ into a prefix block $\mathbf{U}$.
The \emph{Visual Generation Expert} receives a visual history block $\mathbf{V}$, encoded by a frozen Cosmos tokenizer~\citep{agarwal2025cosmos}, and predicts the future visual latent $\mathbf{z}^{v}_{t+15}$.
The \emph{Action Expert} receives the robot state $s_t$ and a noisy action block $\mathbf{A}$, then denoises an action chunk $\mathbf{a}_t$ of length $H=50$ through flow matching~\citep{lipman2022flow}.
Let $\epsilon\sim\mathcal{N}(0,\mathbf{I})$ and $\sigma\in[0,1]$ be the flow-matching time; the noisy chunk is $\mathbf{a}^{\sigma}_t=\sigma\epsilon+(1-\sigma)\mathbf{a}_t$, and the expert predicts a velocity $v_\theta$ toward $\epsilon-\mathbf{a}_t$:
\begin{equation}
\mathcal{L}_{\mathrm{act}}=\mathbb{E}_{\sigma,\epsilon}\left\|v_\theta\!\left(\mathbf{a}^{\sigma}_t,\sigma\mid\mathbf{U},\mathbf{V},s_t\right)-(\epsilon-\mathbf{a}_t)\right\|_2^2 .
\label{eq:act}
\end{equation}
The visual latent prediction loss $\mathcal{L}_{\mathrm{vis}}$ is the mean squared error to the Cosmos latent of $I_{t+15}$, and the base model is trained with $\mathcal{L}_{\mathrm{act}}+\lambda_{\mathrm{vis}}\mathcal{L}_{\mathrm{vis}}$.

\textbf{Problem Setup.}
We extend the base model with $K$ vision-based tactile sensors, two per gripper.
Let $\mathbf{x}^{k}_t$ denote the reference-subtracted RGB observation (Appendix~\ref{app:preprocess}) from sensor $k$ at time $t$, and $\mathbf{x}^{k}_{t-3:t}$ its four-frame history.
Given $\mathcal{C}_t=(\mathbf{U},\mathbf{V},\{\mathbf{x}^{k}_{t-3:t}\}_{k=1}^{K},s_t)$, the model predicts
\begin{equation}
\hat{\mathbf{z}}^{v}_{t+15}=f_{\mathrm{vis}}(\mathcal{C}_t),\qquad
\big(\hat{\mathbf{z}}^{k}_{t+h},\Delta\hat{\mathbf{z}}^{k}_{t+h}\big)=f_{\mathrm{tac}}(\mathcal{C}_t),\qquad
\hat{\mathbf{a}}_t=f_{\mathrm{act}}(\mathcal{C}_t),
\label{eq:setup}
\end{equation}
where $h\in\mathcal{H}=\{5,15,25\}$.
The tactile targets are the future representation $\mathbf{z}^{k}_{t+h}=E(\mathbf{x}^{k}_{t+h})$ produced by a frozen dynamics-aware tactile encoder $E$ (Section~\ref{sec:tacrep}) and its change from the present, $\Delta\mathbf{z}^{k}_{t+h}=\mathbf{z}^{k}_{t+h}-\mathbf{z}^{k}_{t}$.
Both world-model predictions use only observed histories and the current state; future observations serve only as training targets.

\begin{figure}[t]
    \centering
    \includegraphics[width=0.85\linewidth]{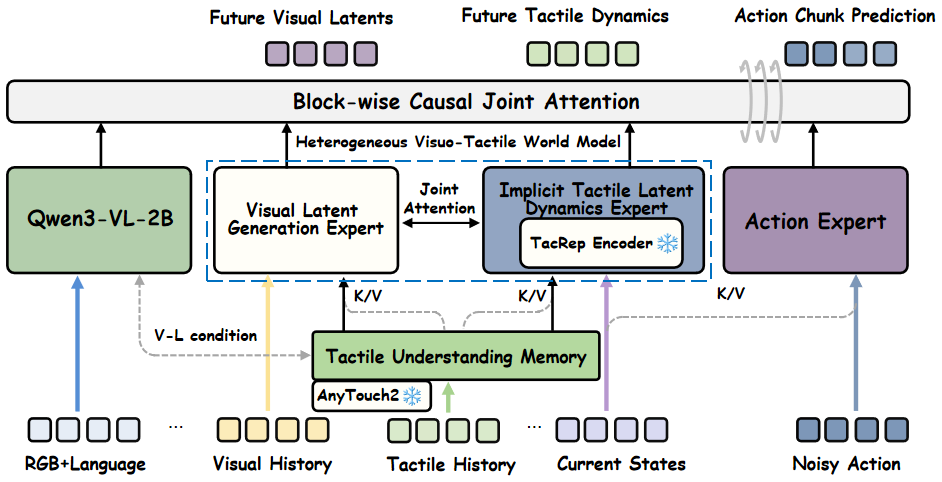}
    \caption{Overview of TacDyn-WAM. The visual and tactile experts form a heterogeneous world model, while the Tactile Understanding Memory provides the current tactile state as read-only keys and values.}
    \label{fig:framework}
\end{figure}

\textbf{TacDyn-WAM Overview.}
TacDyn-WAM adds the \emph{Implicit Tactile Dynamics Expert} (ITDE) as a fourth expert (Figure~\ref{fig:framework}).
The ITDE operates on a tactile block $\mathbf{T}$ encoded by the frozen \emph{TacRep Encoder} $E$ and predicts future tactile representations and their changes at multiple horizons.
Together with the \emph{Visual Latent Generation Expert} of the base model, it forms a \emph{Heterogeneous Visuo-Tactile World Model}: the two experts use separate parameters and target spaces while interacting through joint attention at every layer.

A separate \emph{Tactile Understanding Memory} compresses the same tactile history into prefix-conditioned tokens and exposes them only as keys and values $\mathbf{M}$, avoiding a fifth expert.
The \emph{Action Expert} can therefore combine predicted visual and tactile futures with the current tactile state.
Three stage-specific causal masks gradually open the connections between the tactile modules and pretrained experts, as detailed in Section~\ref{sec:training}.

\subsection{TacRep: A Dynamics-Aware Tactile Representation Space}
\label{sec:tacrep}

\begin{figure}[t]
    \centering
    \includegraphics[width=0.75\linewidth]{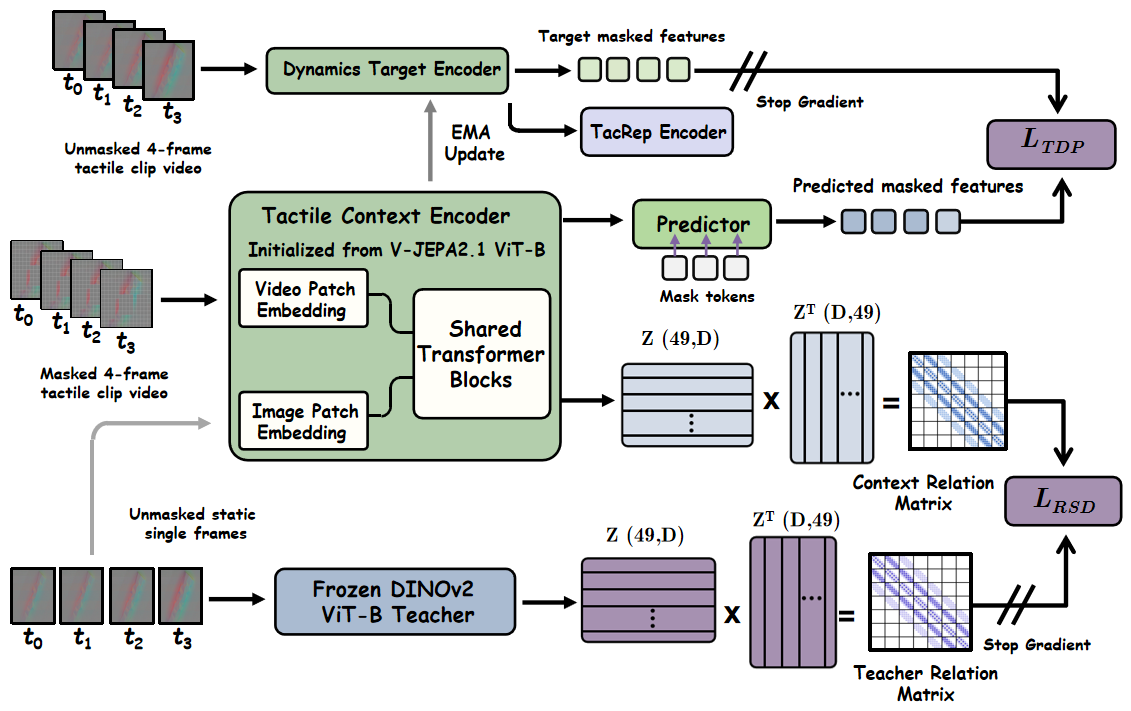}
    \caption{Stage 1: TacRep training. Tactile Dynamics Prediction trains the encoder against an EMA target on masked four-frame clips. Relational Structure Distillation keeps local patch relations close to a frozen DINOv2. Only the EMA encoder is kept.}
    \label{fig:stage1}
\end{figure}

\textbf{Why the Target Space Matters.}
As argued in Section~\ref{sec:intro}, deployment drift can expose a policy to tactile images outside the demonstration distribution.
Their pixels may change substantially even when the underlying physical trend remains broadly consistent.
Reconstructing future pixels or VAE latents ties predictions to these unstable pixel patterns and may therefore produce unreliable forecasts under such shifts.
We instead let the ITDE predict tactile dynamics in a representation space designed to capture contact evolution while tolerating pixel-level deviations.
Following V-JEPA~2.1~\citep{mur2026v}, which learns temporal and physical regularities from video, we adapt its encoder to tactile video clips so that the resulting features capture how an imprint moves and deforms.
A frozen DINOv2~\citep{oquab2023dinov2} then serves as a structural teacher, regularizing local patch relations to support robust prediction when tactile inputs deviate from the demonstrations.

\textbf{Tactile Dynamics Prediction.}
Figure~\ref{fig:stage1} illustrates the training.
The \emph{Tactile Context Encoder} $E_\theta$ is initialized from the distilled V-JEPA~2.1 ViT-B checkpoint.
It takes a four-frame tactile video clip $\mathbf{X}=\mathbf{x}^{k}_{t-3:t}$ and groups the frames into tubelets of two, so a clip becomes a $2\times24\times24$ grid of patch tokens.
The \emph{Dynamics Target Encoder} $\bar{E}$ shares the same architecture and is updated as an exponential moving average of $E_\theta$.
Following V-JEPA~2.1, we mask a set of spatio-temporal blocks $\mathcal{M}$, encode the visible tokens with $E_\theta$, and train a light predictor $P_\phi$ to recover the target-encoder features of the masked tokens:
\begin{equation}
\mathcal{L}_{\text{TDP}}=\mathbb{E}_{\mathcal{M}}\Big\|P_\phi\big(E_\theta(\mathbf{X}\setminus\mathcal{M}),\mathcal{M}\big)-\operatorname{sg}\big[\bar{E}(\mathbf{X})\big]_{\mathcal{M}}\Big\|_1 ,
\label{eq:tdp}
\end{equation}
where $\operatorname{sg}[\cdot]$ denotes stop-gradient.
We use two mask views from V-JEPA~2.1, one with many small blocks and one with few large blocks, and add the same loss on the visible tokens with weight 0.5.
Since every masked block spans all four frames, the encoder can recover it only by understanding how the imprint moves across the clip.
The resulting features therefore respond to the trend of the contact rather than to the static pixel pattern of the gel.

\textbf{Relational Structure Distillation.}
TDP captures temporal evolution but does not explicitly preserve local spatial relations, which may become unstable when tactile inputs deviate from the demonstrations.
We therefore use a frozen DINOv2 ViT-B as a \emph{Relational Structure Teacher} to regularize these relations, while temporal dynamics remain learned through TDP.
For RSD, the same four frames are processed independently as unmasked static tactile images.
The Tactile Context Encoder processes them through its image path with tubelet size one, and the teacher processes them with its own input normalization.
Each frame yields a feature grid that we pool to $7\times7$ and normalize.
Let $\mathbf{z}_i$ and $\mathbf{d}_i$ be the features of patch $i$ from $E_\theta$ and the teacher, respectively.
Since V-JEPA~2.1 and DINOv2 define different latent spaces, directly regressing $\mathbf{z}_i$ to $\mathbf{d}_i$ would create conflicting feature targets, pulling the tactile features toward a static image representation and weakening the dynamics learned by TDP.
Instead, we align their local cosine relations, transferring spatial structure without forcing the feature values to match:
\begin{equation}
\mathcal{L}_{\text{RSD}}=\frac{1}{|\mathcal{N}|}\sum_{(i,j)\in\mathcal{N}}\operatorname{SmoothL1}\big(\mathbf{z}_i^{\top}\mathbf{z}_j-\mathbf{d}_i^{\top}\mathbf{d}_j\big),
\label{eq:rsd}
\end{equation}
where $\mathcal{N}$ contains all patch pairs that lie within two rows and two columns of each other on the $7\times7$ grid.
The Stage 1 objective is $\mathcal{L}_{\text{TDP}}+\lambda_{\text{RSD}}\mathcal{L}_{\text{RSD}}$, with a small $\lambda_{\text{RSD}}$ so that RSD regularizes spatial structure without dominating TDP.
After training, the predictor and teacher are discarded, while the Dynamics Target Encoder $\bar{E}$ is retained as the frozen \emph{TacRep Encoder} $E$ for subsequent ITDE training.

\subsection{Implicit Tactile Dynamics Expert}
\label{sec:itde}

\begin{wrapfigure}{r}{0.5\textwidth}
    \vspace{-10pt}
    \centering
    \includegraphics[width=0.5\textwidth]{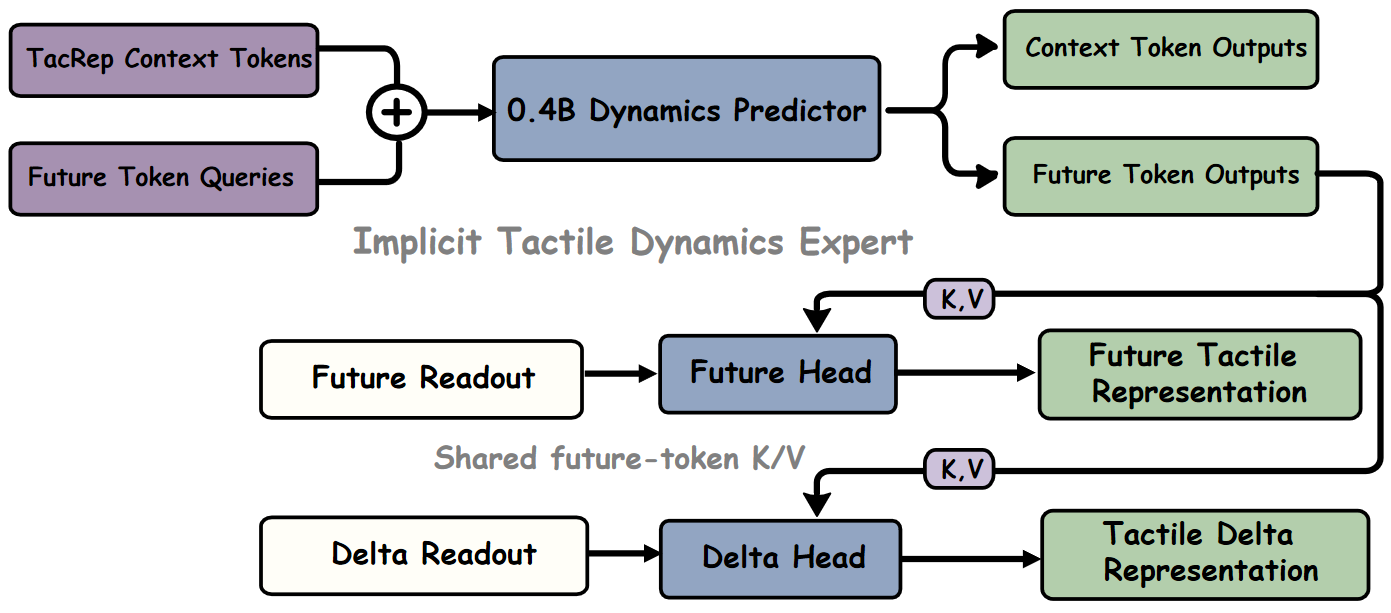}
    \caption{Implicit Tactile Dynamics Expert. A 0.4B dynamics predictor maps context tokens and future queries to the Future and Delta representations at three horizons.}
    \label{fig:itde}
    \vspace{-8pt}
\end{wrapfigure}

\textbf{Context Tokens and Future Queries.}
As shown in Figure~\ref{fig:itde}, the frozen TacRep Encoder converts the four-frame tactile clip from each sensor into two temporal feature maps.
Each map is average-pooled to a $4\times4$ grid and projected to the ITDE hidden dimension, yielding 32 \emph{TacRep Context Tokens} per sensor.
For each of the three prediction horizons $h\in\mathcal{H}=\{5,15,25\}$ frames, we append 16 learnable \emph{Future Token Queries} arranged on a $4\times4$ spatial grid.
Both context and future tokens receive spatial and modality embeddings, together with a continuous time embedding determined by their frame offsets.
The current robot state $s_t$ is injected after every Transformer layer through a zero-initialized conditioning block.

\textbf{Dynamics Predictor.}
The 80 tokens per sensor pass through a 28-layer Transformer with hidden size 1024, about 0.4B parameters.
It has the same layer geometry as the visual and action experts so that all experts can share attention, but it is initialized from scratch.
The outputs of the 32 context tokens and the 48 future tokens together form the tactile block $\mathbf{T}$ that the other experts read.

\textbf{Future and Delta Representations.}
The future tokens are supervised through two independent fine-grained readouts, so the 16 compact tokens of each horizon are expanded to a $7\times7$ prediction outside the 0.4B predictor.
The \emph{Future Readout} and the \emph{Delta Readout} each hold 49 learnable queries on a $7\times7$ grid and cross-attend to the 16 future tokens of one horizon, followed by a linear head to the TacRep dimension.
The targets come from the frozen TacRep Encoder applied to single frames.
For horizon $h$, we encode the current frame and the future frame, pool the feature maps to $7\times7$, and normalize each patch:
\begin{equation}
\mathbf{z}^{k}_{t+h}=\operatorname{norm}\big(\operatorname{pool}(E(\mathbf{x}^{k}_{t+h}))\big),\qquad
\Delta\mathbf{z}^{k}_{t+h}=\mathbf{z}^{k}_{t+h}-\mathbf{z}^{k}_{t}.
\label{eq:targets}
\end{equation}
The Future Readout predicts the \emph{Future Tactile Representation} $\mathbf{z}^{k}_{t+h}$, with its output normalized in the same way, and the Delta Readout predicts the \emph{Tactile Delta Representation} $\Delta\mathbf{z}^{k}_{t+h}$.
Predicting only the future representation may encourage memorization of typical end states, whereas a delta alone is ambiguous without its starting tactile state.
Predicting both requires the ITDE to model how the current tactile state evolves into the future.

\textbf{Patch-Adaptive Loss Reweighting.}
Both readouts are optimized with a Smooth L1 loss.
Since most patches remain nearly unchanged over a prediction horizon, an unweighted average would be dominated by static regions.
We therefore weight the loss of each patch according to the magnitude of its ground-truth change:
\begin{equation}
\begin{aligned}
\mathcal{L}_{\text{tac}}&=\tfrac{1}{2|\mathcal{H}|P}\sum_{h\in\mathcal{H}}\sum_{p}w_{h,p}\Big[\operatorname{SmoothL1}\big(\hat{\mathbf{z}}_{t+h,p}-\mathbf{z}_{t+h,p}\big)+\operatorname{SmoothL1}\big(\Delta\hat{\mathbf{z}}_{t+h,p}-\Delta\mathbf{z}_{t+h,p}\big)\Big],\\
w_{h,p}&\propto 1+\alpha\,\|\Delta\mathbf{z}_{t+h,p}\|_2 .
\end{aligned}
\label{eq:ltac}
\end{equation}
Here, $p$ indexes the $P=49$ patches of each sensor, and the loss is further averaged over sensors.
Within each prediction horizon, the weights are normalized to have a mean of one and capped at a maximum value.
This weighting focuses learning on regions undergoing contact expansion, contraction, or sliding, while retaining basic supervision for static patches.
All predictions are produced in a single forward pass without iterative denoising.
The readouts are used only during training; at deployment, the \emph{Action Expert} directly reads the tactile block $\mathbf{T}$.

\subsection{Tactile Understanding Memory}
\label{sec:memory}

\textbf{Compact Tactile Memory.}
The understanding pathway summarizes the current tactile state: we encode the same four-frame tactile clip with a frozen AnyTouch2 encoder~\citep{feng2026anytouch} pretrained on a large multi-sensor corpus.
Since its output contains several hundred tokens per sensor, the \emph{Compact Attentive Pooler} compresses them into 10 tokens: one global token from the class embedding and nine local tokens obtained by pooling the latest temporal feature map into a $3\times3$ grid.
These tokens are refined through cross-attention over the full AnyTouch2 sequence.
A two-layer \emph{Compact Tactile Q-Former} then fuses tokens across sensors through self-attention and conditions them on the prefix $\mathbf{U}$ through cross-attention.
We use AnyTouch2 only in the understanding pathway: its reconstruction-based pretraining preserves fine-grained tactile information about the current contact, whereas the ITDE operates in the TacRep space designed specifically for predicting future tactile dynamics without pixel reconstruction.

\textbf{Read-Only Integration.}
The memory tokens stay outside the residual streams of all experts.
Each joint-attention layer maps them into additional keys and values through its own lightweight projections, and only the experts permitted by the causal mask can read them.
Because the memory contributes no queries, its tokens are never updated.
They occupy fixed position IDs right after the prefix, add only about 1\% to the parameter count, and leave the original blocks unaffected when masked out.

\subsection{Heterogeneous Joint Attention and Staged Training}
\label{sec:training}

\begin{figure}[t]
    \centering
    \includegraphics[width=0.72\linewidth]{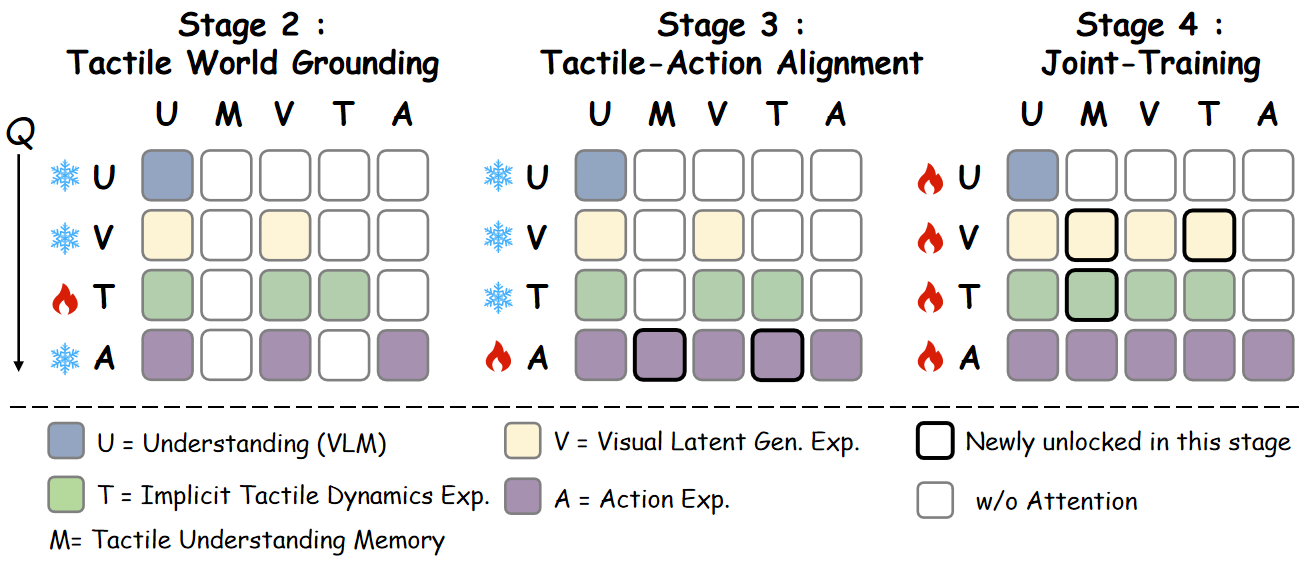}
    \caption{Attention visibility and trainable modules in Stages 2 to 4. Rows are queries and columns are keys; U, M, V, T, and A denote the prefix, memory, visual, tactile, and action blocks. Bold marks connections unlocked in each stage and a flame marks trained modules.}
    \label{fig:mask}
\end{figure}

\textbf{Joint Attention.}
At each layer, the prefix, visual, tactile, and action blocks compute separate queries, keys, and values, which are concatenated and processed by joint attention under a block-wise causal mask (Figure~\ref{fig:mask}).
In the final stage, the prefix attends only to itself; the visual and tactile blocks attend to the prefix, memory, and both world model blocks; and the action block attends to all context and causally to itself.

\textbf{Staged Training.}
Because the ITDE, tactile memory, and readouts are randomly initialized, training the entire model from the outset could disrupt the pretrained experts.
We therefore use four stages and gradually open their attention connections.
\emph{Stage 1: TacRep Training} learns the tactile representation as described in Section~\ref{sec:tacrep} and produces the frozen TacRep Encoder used by all later stages.
\emph{Stage 2: Tactile World Grounding} trains the ITDE and its readouts with $\mathcal{L}_{\text{tac}}$, while the other experts remain frozen; the ITDE reads the prefix and visual block but is not yet visible to them.
\emph{Stage 3: Tactile--Action Alignment} trains the tactile memory and \emph{Action Expert} with $\mathcal{L}_{\text{act}}$, allowing the action block to read both tactile pathways.
\emph{Stage 4: Joint Training} opens the remaining connections and optimizes all trainable modules with
\begin{equation}
\mathcal{L}=\mathcal{L}_{\text{act}}+\lambda_{\text{vis}}\mathcal{L}_{\text{vis}}+\lambda_{\text{tac}}\mathcal{L}_{\text{tac}} .
\label{eq:total}
\end{equation}
Retaining $\mathcal{L}_{\text{tac}}$ preserves the predictive role of the tactile block and prevents it from becoming an action shortcut.
Training details and hyperparameters are provided in Appendix~\ref{app:impl}.

\section{Experiments}
\label{sec:exp}

\subsection{Experimental Setup}
\label{sec:setup}

\textbf{Benchmark.}
UniVTAC~\citep{chen2026univtac} is a challenging simulation benchmark for contact-rich manipulation, featuring a parallel gripper with a visuo-tactile sensor on each finger.
It comprises eight tasks spanning pose reasoning, shape perception, and precise contact interaction (Table~\ref{tab:main}).

\textbf{Baselines.}
Table~\ref{tab:main} compares TacDyn-WAM against vision-only and tactile baselines, including three tactile methods with large-scale tactile pretraining; we report their published UniVTAC results and re-evaluate our base model InternVLA-A1 under our own protocol (Appendix~\ref{app:baselines}).

\textbf{Protocol and Implementation.}
We use the marker-free tactile RGB images from the GelSight Mini sensors on both gripper fingers, together with a head camera and a wrist camera.
Following the benchmark, we train one policy per task on the demonstrations it provides for that task and report the success rate over 100 rollouts per task.
Training details are given in Appendix~\ref{app:impl} and evaluation details in Appendix~\ref{app:eval}.

\subsection{Main Results on UniVTAC}
\label{sec:main}

\begin{table*}[t]
\centering
\setlength{\belowcaptionskip}{4pt}
\caption{Success rates (\%) on the eight UniVTAC tasks. Baselines are grouped into vision-only policies, visuo-tactile policies, and policies with large-scale visuo-tactile trajectory pretraining. Gray rows are excluded from comparison, ``--'' marks results not reported, and the best participating result per column is bolded.}
\label{tab:main}
\scriptsize\setlength{\tabcolsep}{3pt}
\begin{tabular}{clccccccccc}
\toprule
& Method & \shortstack{Insert\\Hole} & \shortstack{Insert\\Tube} & \shortstack{Lift\\Can} & \shortstack{Pull-out\\Key} & \shortstack{Put\\Bottle} & \shortstack{Lift\\Bottle} & \shortstack{Grasp\\Classify} & \shortstack{Insert\\HDMI} & Avg. \\
\midrule
\multirow{7}{*}{\shortstack[c]{Vision-only\\Policies}} & $\pi_{0.5}$~\citep{intelligence2025pi_} & 25 & 74 & 6 & 35 & 34 & \textbf{100} & 49 & 8 & 41.4 \\
& StarVLA-$\alpha$~\citep{ye2026starvla} & 52 & 69 & 65 & 51 & \textbf{88} & 32 & 68 & 24 & 56.1 \\
& InternVLA-A1~\citep{cai2026internvla} & 70 & 56 & 58 & 67 & 37 & 58 & 88 & 12 & 55.8 \\
& Xiaomi-Robotics-0~\citep{cai2026xiaomi} & \textbf{96} & \textbf{98} & 13 & 80 & 12 & 21 & 45 & \textbf{69} & 54.3 \\
& GigaWorld-Policy~\citep{ye2026gigaworld} & 12 & 9 & 0 & 32 & 21 & 38 & 20 & 0 & 16.5 \\
& LingBot-VA~\citep{li2026causal} & 42 & 96 & 0 & 58 & 0 & 0 & 17 & 38 & 31.4 \\
& Fast-WAM~\citep{yuan2026fast} & 66 & 98 & 0 & 73 & 35 & 21 & 72 & 19 & 48.0 \\
\midrule
\multirow{5}{*}{\shortstack[c]{Visuo-tactile\\Policies}} & ACT+UniVTAC~\citep{chen2026univtac} & 24 & 56 & 29 & 46 & 31 & 71 & 99 & 28 & 48.0 \\
& VITaL~\citep{george2025vital} & 25 & 34 & 8 & 47 & 32 & 72 & \textbf{100} & 6 & 40.5 \\
& RDP~\citep{xue2025reactive} & 23 & 75 & 12 & 18 & 41 & 84 & -- & -- & 42.2 \\
& TacForcing~\citep{zhou2026tacforcing} & 69 & 79 & 63 & 48 & 43 & 90 & -- & -- & 65.3 \\
& Tactile-WAM~\citep{wu2026tactilewam} & 20 & 85 & 10 & 20 & 45 & 5 & 55 & 0 & 30.0 \\
\midrule
\rowcolor{white} & \textcolor{gray}{FTP-1~\citep{yuan2026ftp}} & \textcolor{gray}{64} & \textcolor{gray}{79} & \textcolor{gray}{65} & \textcolor{gray}{48} & \textcolor{gray}{47} & \textcolor{gray}{97} & \textcolor{gray}{99} & \textcolor{gray}{4} & \textcolor{gray}{62.9} \\
\rowcolor{white}& \textcolor{gray}{$N_0$-VTLA~\citep{neoteai2026n0vtla}} & \textcolor{gray}{95} & \textcolor{gray}{99} & \textcolor{gray}{88} & \textcolor{gray}{99} & \textcolor{gray}{60} & \textcolor{gray}{99} & \textcolor{gray}{100} & \textcolor{gray}{25} & \textcolor{gray}{83.1} \\
\rowcolor{white}& \textcolor{gray}{$N_0$-TWAM (20\% data)} & \textcolor{gray}{78} & \textcolor{gray}{96} & \textcolor{gray}{80} & \textcolor{gray}{25} & \textcolor{gray}{46} & \textcolor{gray}{51} & \textcolor{gray}{95} & \textcolor{gray}{52} & \textcolor{gray}{65.4} \\
\rowcolor{white}\textcolor{gray}{\multirow{-4}{*}{\shortstack[c]{Visuo-tactile\\Pretrained\\Policies}}} & \textcolor{gray}{$N_0$-TWAM~\citep{neoteai2026n0twam}} & \textcolor{gray}{99} & \textcolor{gray}{98} & \textcolor{gray}{93} & \textcolor{gray}{79} & \textcolor{gray}{87} & \textcolor{gray}{58} & \textcolor{gray}{94} & \textcolor{gray}{68} & \textcolor{gray}{84.5} \\
\midrule
& TacDyn-WAM (Ours) & 91 & 97 & \textbf{87} & \textbf{97} & 72 & 97 & 99 & 12 & \textbf{81.5} \\
\bottomrule
\end{tabular}
\end{table*}

Table~\ref{tab:main} reports the results.
TacDyn-WAM reaches an average success rate of 81.5\%, second only to the two $N_0$ models, and is within 3 points of the best baseline on Insert Tube, Pull-out Key, Lift Bottle, and Grasp Classify.

\textbf{Comparison with Vision-Only Models.}
TacDyn-WAM improves its base model InternVLA-A1 by 25.7 points and exceeds StarVLA-$\alpha$, the strongest vision-only VLA, by 25.4 points.
The vision-only world action models fall further behind, with the best of them, Fast-WAM, at 48.0\%, and TacDyn-WAM reaches at least 72\% on seven of the eight tasks.

\textbf{Comparison with Tactile Models.}
The UniVTAC policy and VITaL, which add a tactile encoder to ACT, reach 48.0\% and 40.5\%; RDP and TacForcing add execution-time tactile adaptation and reach 42.2\% and 65.3\%, while Tactile-WAM, with tactile asymmetric attention, reaches 30.0\%.
FTP-1 is pretrained on about 3,000 hours of tactile data, yet stays 18.6 points below TacDyn-WAM.
$N_0$-VTLA and $N_0$-TWAM are pretrained on more than 30,000 hours of visuo-tactile trajectories from six embodiments and 450 tasks, and $N_0$-TWAM has 7.2B parameters.
TacDyn-WAM has about 3.6B parameters and is trained on the per-task demonstrations alone, yet it is only 1.6 and 3.0 points behind them, and it beats $N_0$-TWAM on Pull-out Key, Lift Bottle, and Grasp Classify.
When $N_0$-TWAM is pretrained on 20\% of its corpus, its average drops to 65.4\%, 16.1 points below ours.
The remaining gaps to $N_0$-TWAM are on Insert Hole and Put Bottle, where its 20\% variant also drops sharply, and on Insert HDMI, whose short and nearly perfect demonstrations contain almost no tactile variation, so success depends on vision and $N_0$-TWAM likely benefits from its fine-grained video prediction.

\subsection{Ablation Study}
\label{sec:ablation}

Table~\ref{tab:ablation} evaluates the contributions of the two tactile pathways through separate ablations and compares TacRep with two frozen alternatives: the base model's Cosmos VAE and a single-frame DINOv2 ViT-B encoder.

\begin{table*}[t]
\centering
\setlength{\belowcaptionskip}{4pt}
\caption{Ablation study on UniVTAC (success rate, \%). All variants share the same base model, data, and per-stage training hyperparameters. More details are given in Appendix~\ref{app:ablation-impl}.}
\label{tab:ablation}
\scriptsize\setlength{\tabcolsep}{4pt}
\begin{tabular}{lccccccccc}
\toprule
Method & \shortstack{Insert\\Hole} & \shortstack{Insert\\Tube} & \shortstack{Lift\\Can} & \shortstack{Pull-out\\Key} & \shortstack{Put\\Bottle} & \shortstack{Lift\\Bottle} & \shortstack{Grasp\\Classify} & \shortstack{Insert\\HDMI} & Avg. \\
\midrule
w/o Tactile World Model & 79 & 72 & 70 & 88 & 52 & 76 & 95 & 11 & 67.9 \\
w/o Tactile Understanding Memory & 84 & 83 & 77 & 82 & 58 & 86 & 91 & 9 & 71.3 \\
TacRep $\rightarrow$ Cosmos VAE & 74 & 69 & 63 & 61 & 47 & 70 & 97 & 15 & 62.0 \\
TacRep $\rightarrow$ DINOv2 & 85 & 83 & 80 & 89 & 61 & 84 & 98 & 12 & 74.0 \\
\midrule
\textbf{TacDyn-WAM (full)} & \textbf{91} & \textbf{97} & \textbf{87} & \textbf{97} & \textbf{72} & \textbf{97} & \textbf{99} & \textbf{12} & \textbf{81.5} \\
\bottomrule
\end{tabular}
\end{table*}

\textbf{Target Space.}
The Cosmos VAE performs worst at 62.0\%, falling 5.9 points below the variant without tactile prediction.
The largest drop occurs on Pull-out Key (61.0\%), where the tactile observations in failed rollouts differ clearly from those in the demonstrations (Appendix~\ref{app:drift}).
Predicting in a pixel-reconstruction space therefore provides unreliable guidance once execution deviates from the demonstrations.
DINOv2 recovers the average to 74.0\% but remains 7.5 points below the full model, with the largest gaps on Insert Tube, Put Bottle, and Lift Bottle, which involve sustained rotation of the contact pattern.
Static features are less tied to exact pixel patterns but lack the temporal sensitivity provided by TacRep.

\textbf{Two Tactile Pathways.}
Removing the tactile world model or memory reduces the average success rate by 13.6 and 10.2 points, respectively, confirming that both pathways contribute.
Tactile prediction provides the largest gains on Insert Tube, Put Bottle, and Lift Bottle, where anticipating contact rotation is important.
The memory instead supplies direct cues about the current tactile state, notably for Pull-out Key and Grasp Classify.

\textbf{When Tactile Modeling Offers Limited Benefit.}
On Insert HDMI, all variants remain within three points of the full model, since its demonstrations contain little tactile variation (Section~\ref{sec:main}).
Grasp Classify is mainly a perception task, so only removing the Tactile Understanding Memory causes a clear drop (91\%).

\subsection{Modest-Scale Tactile Pretraining}
\label{sec:pretrain}

We further pretrain TacDyn-WAM on a modest amount of real visuo-tactile data as the starting point for the real-robot experiments in Section~\ref{sec:real}.

\textbf{Data.}
OmniViTac~\citep{zheng2026omnivta} is a real-world visuo-tactile dataset of 21,879 trajectories over 86 tasks, collected with an xArm-7 and a handheld TacUMI device with the same parallel-jaw gripper.
Since our real-robot tasks are mostly related to assembly, we select a subset of 6,000 trajectories: 2,500 from assembly, 2,000 from grasping, 1,000 from wiping, and 500 from adjustment, leaving out cutting and peeling.
We also include the 300 demonstrations that we collected for the five real-robot tasks in Section~\ref{sec:real}, 60 per task.
Both sources use the same Xense vision-based tactile sensor on each finger, and we directly use its output tactile difference images, which already remove the no-contact background.

\textbf{Procedure.}
Pretraining follows the four stages of Section~\ref{sec:training}, with the two data sources entering at different stages.
Stages 1 and 2 train the TacRep Encoder and the ITDE without action supervision.
We include the real-robot demonstrations in these two stages, with oversampling, so that the tactile representation and dynamics prediction already cover the contacts of the five real tasks before fine-tuning.
Stages 3 and 4 require action supervision and use OmniViTac only.
Although its end-effector actions differ from the joint actions of our robot, these stages let the experts learn to work together on real contact data: the Action Expert learns to read the tactile block and the Tactile Understanding Memory, and the tactile modules adapt jointly with the pretrained experts.
The two action spaces occupy separate slots of a shared action vector, and after pretraining the model is fine-tuned per task with Stage 4 on its 60 real-robot demonstrations.
Data statistics and hyperparameters are listed in Appendix~\ref{app:impl-pretrain}.

\subsection{Real-World Experiments}
\label{sec:real}

\textbf{Setup.}
We use a Franka Research 3 arm with a parallel gripper, one Xense vision-based tactile sensor on each finger, and a wrist camera and a third-person camera.
We collect 60 teleoperated demonstrations for each of five contact-rich tasks: Stack Cups, Remove Plug, Insert Plug, Unscrew Cup Lid, and Wipe Whiteboard (Appendix~\ref{app:real-tasks}).
We compare with two vision-only world action models, LingBot-VA and InternVLA-A1, and with FTP-1, a tactile policy pretrained on about 3,000 hours of tactile data, all fine-tuned per task on the same demonstrations.
TacDyn-WAM is evaluated in two variants: trained per task through the four stages as in simulation, or pretrained as in Section~\ref{sec:pretrain} and then fine-tuned per task with Stage 4.
Each policy runs 20 trials per task from randomized initial configurations (Appendix~\ref{app:eval-real}).

\begin{table}[t]
\centering
\setlength{\belowcaptionskip}{4pt}
\caption{Real-world success rates (\%) over 20 trials per task. All baselines are fine-tuned per task on the same demonstrations.}
\label{tab:real}
\scriptsize\setlength{\tabcolsep}{5pt}
\begin{tabular}{lcccccc}
\toprule
Method & \shortstack{Stack\\Cups} & \shortstack{Remove\\Plug} & \shortstack{Insert\\Plug} & \shortstack{Unscrew\\Cup Lid} & \shortstack{Wipe\\Whiteboard} & Avg. \\
\midrule
LingBot-VA & 40 & 10 & 45 & 15 & 50 & 32.0 \\
InternVLA-A1 & 55 & 20 & 45 & 10 & 70 & 40.0 \\
FTP-1 & 65 & 45 & 65 & 20 & 80 & 55.0 \\
\midrule
TacDyn-WAM & 80 & 60 & 80 & 40 & 95 & 71.0 \\
TacDyn-WAM (pretrained) & \textbf{95} & \textbf{85} & \textbf{90} & \textbf{55} & \textbf{100} & \textbf{85.0} \\
\bottomrule
\end{tabular}
\end{table}

\begin{table}[t]
\centering
\setlength{\belowcaptionskip}{4pt}
\caption{Inference latency on one A100 GPU with batch size 1. LingBot-VA (naive) recomputes the history at every step; LingBot-VA (optimized) uses the deployment settings of its paper.}
\label{tab:speed}
\scriptsize\setlength{\tabcolsep}{5pt}
\begin{tabular}{lccc}
\toprule
Method & Actions per chunk & Latency per chunk (ms) & Latency per action (ms) \\
\midrule
LingBot-VA (naive) & 40 & 2,211 & 55.3 \\
LingBot-VA (optimized) & 40 & 1,338 & 33.5 \\
TacDyn-WAM (ours) & 50 & \textbf{484} & \textbf{9.7} \\
\bottomrule
\end{tabular}
\end{table}

\textbf{Results.}
Without the OmniViTac pretraining stage, TacDyn-WAM reaches 71.0\% (Table~\ref{tab:real}), 16 points above FTP-1 and 31 points above its base model InternVLA-A1.
The gains are largest on Remove Plug and Unscrew Cup Lid, where the vision-only policies often slip during pulling or tip the bottle over when the lid is not turned to the right position.
On Stack Cups, vision-only policies tend to keep pushing downward in a wrong direction, whereas TacDyn-WAM lifts the cup and re-aligns it when the contact does not match the expected evolution.
Modest-scale pretraining improves every task and raises the average to 85.0\%, with the largest gains on Remove Plug and Stack Cups, while Unscrew Cup Lid remains the hardest task for all methods.

\textbf{Inference Latency.}
Table~\ref{tab:speed} compares TacDyn-WAM with LingBot-VA, which predicts future video frames by iterative denoising, under the same GPU and synchronous setting (Appendix~\ref{app:eval-real}).
Even with the deployment optimizations of its paper, LingBot-VA needs 2.8 times the latency of TacDyn-WAM for a shorter chunk, and its latency grows as the episode history accumulates.
TacDyn-WAM predicts in compact latent spaces without iterative video denoising, so its cost is lower and constant over an episode.

\section{Conclusion and Limitations}
\label{sec:conclusion}

We presented TacDyn-WAM, a heterogeneous visuo-tactile world action model that predicts future visual latents and tactile dynamics in separate target spaces and couples them through joint attention.
On UniVTAC, TacDyn-WAM reaches state-of-the-art-level performance with per-task demonstrations alone, and modest-scale pretraining further improves real-robot manipulation.
One limitation is that TacRep and the ITDE are trained for a single type of visuo-tactile sensor; generalizing them across tactile sensors, including non-image modalities such as force and taxel arrays, is left for future work.

\subsection*{AI Use Statement}
We used generative AI tools to aid and polish the writing of this paper.
They also assisted with writing part of the implementation code under the authors' direction.
The research questions, method design, core implementation, experiments, analyses, and conclusions are the authors' own, and the authors take full responsibility for the content of the paper.

\subsection*{Reproducibility Statement}
Several measures have been taken to make our work reproducible.
The main paper describes the model architecture, training objectives, staged training procedure, and evaluation setup.
The appendix provides the input preprocessing, per-stage hyperparameters, ablation and baseline details, and the evaluation protocol.
Simulation experiments are conducted on the public UniVTAC benchmark with publicly available pretrained models.
For the real-robot experiments, the appendix provides the task definitions, data-collection procedure, and evaluation criteria.
Related demo and core code are provided as supplementary material.

\bibliography{iclr2027_conference}
\bibliographystyle{iclr2027_conference}

\appendix

\section{Related Work}
\label{app:related}

\subsection{Vision--Tactile--Language--Action Models}

Vision-Language-Action (VLA) models such as the $\pi$ series transfer large-scale vision-language pretraining to robot control~\citep{black2024pi0,intelligence2025pi_,cai2026internvla}.
Because vision alone cannot fully resolve local contact geometry, stability, or interaction forces, vision--tactile--language--action (VTLA) models additionally condition policies on tactile observations.
Representative approaches include Tactile-VLA, OmniVTLA, and $N_0$-VTLA, and further work extends this line to tactile-language control, contact-aware fusion, and tactile--force alignment~\citep{huang2025tactilevla,cheng2025omnivtla,neoteai2026n0vtla,hao2025tla,zhang2025vtla,zhang2026tacvla,huang2026tactile,jones2025beyond}.
Despite their different fusion mechanisms, these methods use tactile signals mainly for perception, where the policy reacts to the current or recent tactile observation.
VLA-Touch further introduces tactile feedback at both the VLM and action-refinement levels, but likewise does not explicitly model future tactile evolution~\citep{bi2025vlatouch}.
In contrast, TacDyn-WAM combines its current-state Tactile Understanding Memory with an explicit predictive tactile dynamics pathway.

A complementary line uses tactile or force signals for reactive action refinement,
ranging from slow--fast policies such as Reactive Diffusion Policy and OmniVTA
to residual correction
\citep{xue2025reactive,zheng2026omnivta,bi2025vlatouch,zhou2026touchworld,he2026fawam,wang2026phaforce,zhou2026tacforcing}.
These methods target rapid feedback after contact, whereas TacDyn-WAM models tactile dynamics inside the world model rather than in an execution-time correction stage.

\subsection{Tactile Representation Learning}

Tactile representations are commonly learned through reconstruction or alignment across sensors and modalities \citep{shan2026learning}.
Reconstruction and masked-modeling methods preserve local detail, whereas alignment across sensors and modalities improves transfer and semantic grounding;
representative examples include Sparsh, M3L, UniT, AnyTouch, and AnyTouch 2
\citep{higuera2024sparsh,sferrazza2024power,xu2025unit,feng2025anytouch,feng2026anytouch}.
Subsequent work extends these ideas to visuo-tactile pretraining, language-grounded
fusion, and generalist policies across tactile sensors
\citep{george2025vital,jones2025beyond,tian2026vitas,yuan2026ftp}. However, reconstruction ties features to exact sensor pixel patterns, and semantic alignment can discard subtle temporal cues such as slip, compression, or contact rotation. ViTacFormer predicts future touch as an
auxiliary policy objective \citep{heng2025vitacformer}, but explicit predictive
representation learning remains less explored. Inspired by the joint-embedding predictive architecture (JEPA) paradigm~\citep{assran2025v,mur2026v}, TacRep instead learns a target space from tactile clips that
emphasizes predictable contact evolution without reconstructing future pixels.
It further regularizes local patch relations with a structural prior, so that the spatial structure of the contact pattern remains stable when tactile inputs deviate from the demonstrations.

\subsection{Tactile World Action Models}

World action models (WAMs) couple future-state prediction with action generation.
Existing WAMs predominantly model visual or semantic dynamics~\citep{yuan2026fast,ye2026world,bi2026motus,li2026causal,wang2026st}.
Recent tactile WAMs can be distinguished by their prediction targets and inference procedures.

\paragraph{Generative tactile prediction.}
A prominent family reuses video-generation pipelines, such as $N_0$-TWAM, Dream-Tac, VTAM, and Tactile-WAM: they encode tactile images with the same video VAE as camera frames and generate future tactile observations through iterative denoising~\citep{neoteai2026n0twam,lou2026dreamtac,yuan2026vtam,wu2026tactilewam,huang2026vitacworld,zhou2026touchworld,jin2026tacwam}.
OmniVTA and VT-WAM instead learn a tactile-specific VAE before generative prediction~\citep{zheng2026omnivta,tian2026vt}.
Although these designs retain fine spatial detail, their targets remain reconstruction-oriented and inference requires multiple denoising steps.
Moreover, pixel similarity does not always track physical similarity: small tactile-image changes may indicate critical slip or rotation, whereas larger pixel changes may be less relevant to control.

\paragraph{Feature and low-dimensional prediction.}
TacForeSight and ContactWorld avoid image generation by predicting compact tactile features in one forward pass~\citep{zang2026tacforesight,zhang2026contactworld}, and $N_0$-VTLA predicts the net change of frozen DINOv2 features over a single horizon~\citep{neoteai2026n0vtla}.
Force-aware WAMs instead predict numeric interaction signals, such as wrist wrench in FAWAM or finger torque in TACO~\citep{he2026fawam,liu2026taco}.
These targets reduce prediction cost, but their utility depends on whether the representation preserves the spatial and temporal contact structure needed for control.
These approaches shift the central design question from how accurately tactile pixels can be reconstructed to which latent dynamics are useful for choosing actions.

TacDyn-WAM follows the one-pass predictive direction, but predicts in a target space learned for tactile dynamics rather than a frozen or reconstruction-oriented one, and keeps this space separate from the visual latent space of the base model.

\subsection{Tactile Fusion Mechanisms}
\label{app:related-fusion}

Tactile fusion mechanisms can be broadly grouped into attention conditioning,
contact-dependent gating, and expert-structured fusion
\citep{shan2026learning}. Token-based VTLA models typically insert tactile
features into a shared sequence or expose them through self- or cross-attention;
Tactile-VLA and OmniVTLA are representative examples
\citep{huang2025tactilevla,cheng2025omnivtla,hao2025tla,zhang2025vtla,neoteai2026n0vtla}.
This design is expressive, but unrestricted fusion can allow a strong
vision-language pathway to dominate optimization or let noisy free-space touch
interfere with visual features.

Gated methods make tactile influence dependent on contact state. TacVLA and
ReTac-ACT learn gates conditioned on the contact or robot state, while OmniVTA uses a lightweight
module to fuse the difference between predicted and observed tactile features
\citep{zhang2026tacvla,ruan2026retac,zheng2026omnivta}. Similar principles
appear as contact-derived attention biases or asymmetric masks in tactile WAMs
\citep{lou2026dreamtac,wu2026tactilewam}. Expert-structured methods instead
preserve modality-specific parameters: ForceVLA uses a force-aware MoE, while
$N_0$-TWAM and VT-WAM separate visual, tactile, and action experts
\citep{yu2026forcevla,neoteai2026n0twam,tian2026vt}.
Gating and expert separation solve related but different problems: a gate
controls the contribution of touch across interaction phases~\citep{wang2026adm}, whereas expert
separation limits parameter interference between heterogeneous modalities.

TacDyn-WAM likewise specializes parameters by modality, but does not rely on a
learned sparse router. Its experts communicate through block-wise joint
attention under stage-specific causal masks. In parallel, the Tactile
Understanding Memory enters only as read-only keys and values rather than as an
additional expert. This design separates current contact
understanding, future contact prediction, and action generation, while staged
training prevents randomly initialized tactile modules from immediately
disrupting the pretrained experts.

\section{Examples of Tactile Distribution Shift between Demonstrations and Policy Execution}
\label{app:drift}

\begin{figure}[htbp]
    \centering
    \begin{minipage}[t]{0.49\linewidth}\centering
        \includegraphics[width=\linewidth]{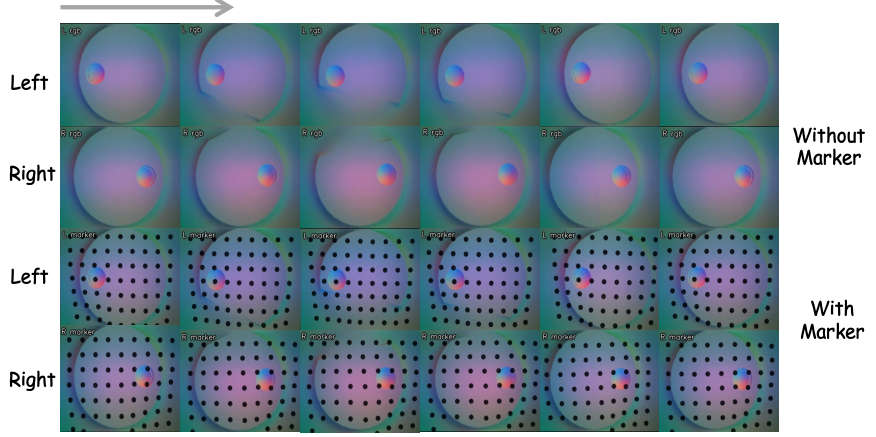}\\[1pt]
        {\small (a) Pull-out Key, expert demonstration}\\[5pt]
        \includegraphics[width=\linewidth]{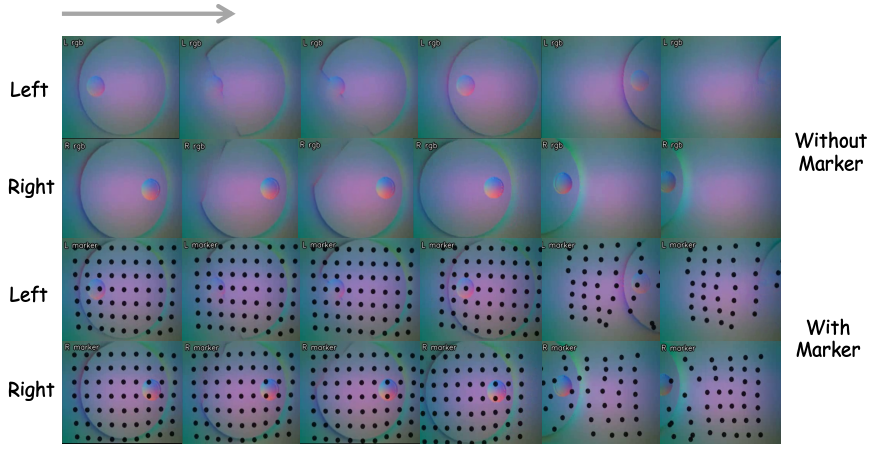}\\[1pt]
        {\small (c) Pull-out Key, policy execution}
    \end{minipage}\hfill
    \begin{minipage}[t]{0.49\linewidth}\centering
        \includegraphics[width=\linewidth]{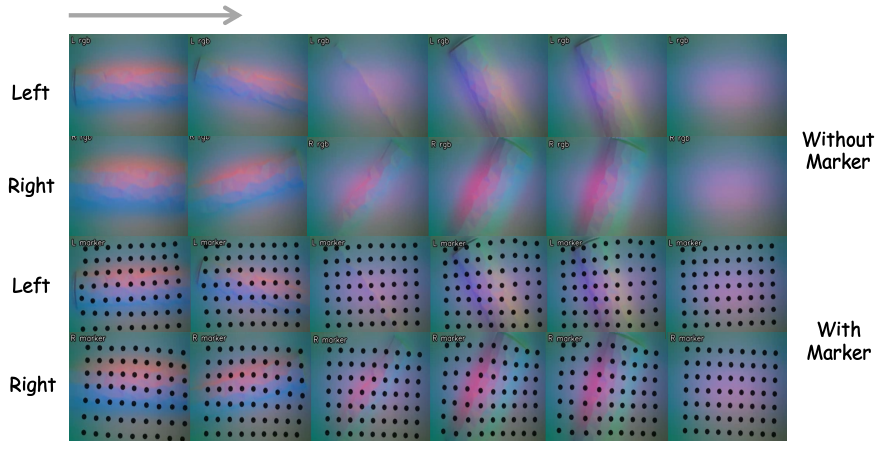}\\[1pt]
        {\small (b) Put Bottle, expert demonstration}\\[5pt]
        \includegraphics[width=\linewidth]{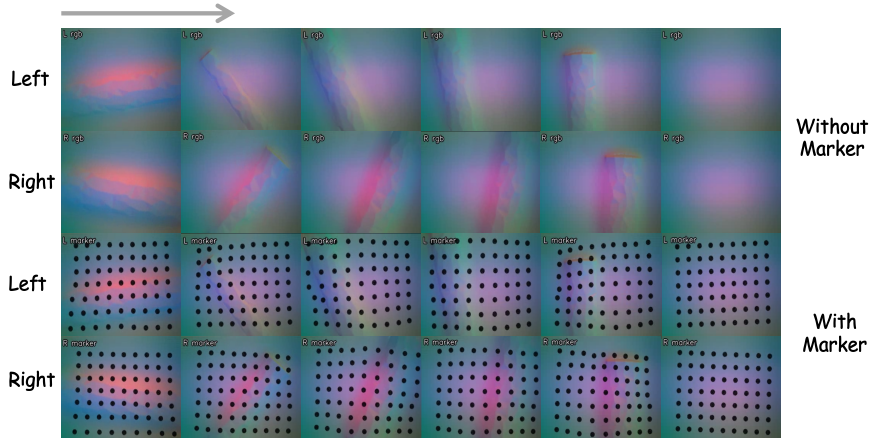}\\[1pt]
        {\small (d) Put Bottle, policy execution}
    \end{minipage}
    \caption{Examples of tactile distribution shift between expert demonstrations (top) and policy execution (bottom) on Pull-out Key (left) and Put Bottle (right). In each panel, rows show the left and right sensors without and with markers, and columns are six frames over time. The imprint follows a similar trend in both cases, but its position, extent, and timing differ during execution, and the marker view makes the difference more visible.}
    \label{fig:drift}
\end{figure}

As examples, Figure~\ref{fig:drift} compares the tactile observations of one expert demonstration with those of one policy execution on Pull-out Key and Put Bottle.
Each panel shows six frames from both sensors of the gripper, in both the marker-free and the marker view of the sensor.
On Pull-out Key, the robot first rotates the key to the unlocking angle and then pulls it out, so the imprint changes clearly during the rotation phase and mostly jitters sideways before and after it.
On Put Bottle, the robot lifts the bottle, rotates it to a suitable angle, and places it on a higher shelf, so the imprint rotates through a sequence of orientations and disappears once the bottle is released.
During execution, the pixel pattern departs from the demonstrations: the imprint is off-center, the final rotation angle differs, and the bottle is sometimes released earlier.
Yet the overall trend is preserved: the imprint rotates in the same direction as in the demonstration.
This is the situation that motivates our target space in Section~\ref{sec:tacrep}: a reconstruction target treats these frames as unseen pixel patterns, while the underlying contact evolution remains predictable.
The marker view makes this shift more visible.
The marker array deforms with the gel, so the same deviation also displaces every marker near the contact.
On Pull-out Key, the later frames of execution show severe marker distortion: the gripper holds the key tightly and pulls upward while the key remains stuck in the lock, so the gel is stretched and torn instead of following the key.
Such deformations never appear in the demonstrations, and they make future tactile images harder to predict and cost extra capacity for a model that reconstructs them.

\section{Tactile Input Preprocessing}
\label{app:preprocess}

\begin{figure}[htbp]
    \centering
    \includegraphics[width=0.65\linewidth]{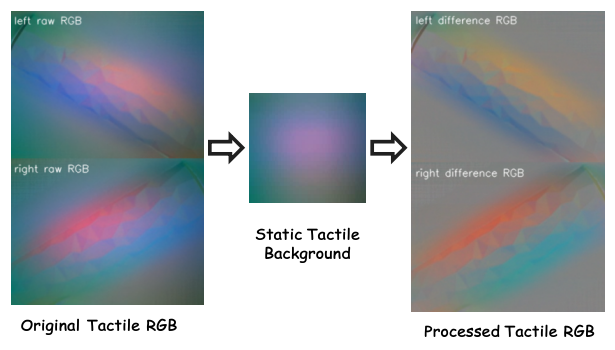}
    \caption{Tactile input preprocessing. Left: raw RGB frames from the two sensors of a gripper. Middle: the static no-contact background frame. Right: the signed residual after subtracting the background and adding a mid-gray offset. Static gel texture and illumination are removed, and only the current deformation remains.}
    \label{fig:bg}
\end{figure}

All tactile inputs to the TacRep Encoder and to AnyTouch2 are background-subtracted images rather than raw sensor frames (Figure~\ref{fig:bg}).
For each sensor we store one static no-contact frame $\mathbf{b}^{k}$, captured once with the gripper open, and share it across all tasks.
A raw frame $\mathbf{r}^{k}_t$ is converted to the signed residual
\begin{equation}
\mathbf{x}^{k}_t=\operatorname{clip}\!\left(\frac{\mathbf{r}^{k}_t-\mathbf{b}^{k}}{255}+\frac{130}{255},\,0,\,1\right),
\label{eq:bg}
\end{equation}
so that a pixel equal to the background maps to a constant mid-gray and positive and negative deformations map to brighter and darker values.
The subtraction is applied at the native sensor resolution before any resizing, and the same offset is used for both tactile pathways.
Raw tactile images contain a fixed gel texture, an illumination gradient, and lens vignetting that are identical in every frame and carry no information about contact.
Subtracting the background removes these static patterns, flattens the non-contact area to a uniform value, and leaves only the deformation caused by the current contact, which makes the input closer to what the model should attend to and reduces the patches that the TacRep Encoder must learn to ignore.

\section{Implementation Details}
\label{app:impl}

All stages are trained in bfloat16 with AdamW, a linear warmup followed by cosine decay, and gradient clipping at a norm of 1.0 applied per parameter group.
The per-stage training settings of the UniVTAC experiments follow in Appendix~\ref{app:impl-sim}.

\subsection{UniVTAC Simulation Experiments}
\label{app:impl-sim}

\textbf{Inputs and Actions.}
At each step the model receives the head and wrist camera frames at $t-15$ and $t$, the four most recent tactile RGB frames from both sensors after background subtraction (Appendix~\ref{app:preprocess}), together with the robot state and task instruction.
Camera frames are 270$\times$480 and tactile RGB frames are 240$\times$320.
The model outputs a chunk of 50 actions, with arm joints expressed relative to the current joint angles and an absolute gripper command.

\textbf{Stage 1.}
The Tactile Context Encoder is initialized from the distilled V-JEPA~2.1 ViT-B checkpoint with 384$\times$384 inputs.
It keeps the two tokenizers of V-JEPA~2.1: the video path embeds a four-frame clip with tubelets of two frames into a $2\times24\times24$ token grid for TDP, and the image path embeds a single frame with a tubelet of one into a $24\times24$ grid for RSD; the two paths use separate patch and modality embeddings but share all Transformer blocks.
The Relational Structure Teacher is a frozen DINOv2 ViT-B/14 with 392$\times$392 inputs, whose $28\times28$ patch grid and the student's $24\times24$ grid are both average-pooled to $7\times7$ before the relation matrices in Equation~\ref{eq:rsd} are computed; $\lambda_{\text{RSD}}=0.1$.
We use the two multiblock mask views of V-JEPA~2.1, a small view with 8 blocks covering 15\% of the spatial area and a large view with 2 blocks covering 70\%, both spanning all four frames, with a context loss weight of 0.5.
To protect the pretrained representation, the encoder is unfrozen gradually: the predictor and the last four Transformer blocks are trained from the start, blocks 4 to 7 are unfrozen after the first 20\% of steps at a lower learning rate, and the first four blocks and the patch embeddings stay frozen throughout.
The EMA momentum of the Dynamics Target Encoder increases linearly from 0.999 to 0.9999.

\textbf{Stage 2.}
The Future and Delta losses are Smooth L1 with $\beta=1$, averaged over the feature dimension per patch and combined as in Equation~\ref{eq:ltac}.
The Patch-Adaptive weights use $\alpha=2$, are normalized to a mean of one within each sensor and horizon over the batch and the 49 patches, and are clipped at 5 so that no single patch dominates the loss.

\textbf{Stage 3.}
The Compact Attentive Pooler, the Compact Tactile Q-Former, and the memory K/V projections use a learning rate of $10^{-4}$, and the Action Expert with its state and action projections $5\times10^{-5}$.

\textbf{Stage 4.}
The loss weights are $\lambda_{\text{vis}}=0.01$ and $\lambda_{\text{tac}}=500$, where $\lambda_{\text{tac}}$ was set by matching the gradient magnitude of the tactile loss to that of the action loss.
Parameters inherited from the base model peak at $5\times10^{-5}$ and decay to $5\times10^{-6}$; the newly added modules peak at $10^{-4}$ and decay to $10^{-5}$.

\textbf{Training Schedule.}
Table~\ref{tab:hparams} lists the optimization settings of the four stages on UniVTAC.
The number of Stage 2 and Stage 3 epochs is scaled per task so that tasks with shorter episodes see a comparable number of optimizer steps.

\begin{table}[htbp]
\centering
\setlength{\belowcaptionskip}{4pt}
\caption{Optimization settings of the four stages on UniVTAC.}
\label{tab:hparams}
\small
\begin{tabular}{lcccc}
\toprule
 & Stage 1 & Stage 2 & Stage 3 & Stage 4 \\
\midrule
Batch size & 64 & 128 & 128 & 64 \\
Peak learning rate & $10^{-4}$ / $10^{-5}$ / $5\times10^{-6}$ & $10^{-4}$ & $10^{-4}$ / $5\times10^{-5}$ & $10^{-4}$ / $5\times10^{-5}$ \\
Weight decay & 0.04 & 0.05 & 0.01 & 0.01 \\
Warmup & 5\% & 5\% & 5\% & 2,000 steps \\
Length & 20 epochs & 15 epochs & 5 epochs & 12,000 steps \\
\bottomrule
\end{tabular}
\end{table}

\subsection{Modest-Scale Tactile Pretraining}
\label{app:impl-pretrain}

\textbf{Data.}
OmniViTac~\citep{zheng2026omnivta} is collected on two embodiments with identical parallel-jaw grippers: an xArm-7 driven by kinesthetic teaching or GELLO teleoperation, and a handheld TacUMI device.
We do not distinguish the two.
Our subset contains 6,000 trajectories: 2,500 assembly, 2,000 grasping, 1,000 wiping, and 500 adjustment, all recorded with Xense sensors.
The real-robot part consists of the 300 demonstrations of the five tasks in Section~\ref{sec:real}, 60 per task, recorded with the same type of Xense sensor.

\textbf{Tactile Input Protocol.}
The Xense sensor outputs tactile difference images with the no-contact background already removed, and both OmniViTac and our real-robot data store this output.
We only shift its zero point to the mid-gray offset of $130/255$ used in Equation~\ref{eq:bg}, by subtracting the per-frame median and adding the offset, so that the input has the same form as the simulation residuals.
The downstream input protocols are unchanged: AnyTouch2 receives 224$\times$224 inputs with CLIP normalization, and the TacRep Encoder receives 384$\times$384 inputs with ImageNet normalization.

\textbf{Action Space.}
The two embodiments occupy non-overlapping slots of a 32-dimensional action vector.
Our robot uses joint actions, seven joint deltas and two absolute gripper values in dimensions 0 to 8.
OmniViTac uses end-effector actions, the position and 6D rotation relative to the current pose plus an absolute gripper value in dimensions 9 to 18.
Each embodiment is normalized with its own statistics, and a validity mask restricts the action loss to the dimensions present in each sample.

\textbf{Schedule.}
Table~\ref{tab:hparams-pretrain} lists the data and training length of each stage.
Within OmniViTac, frames are sampled uniformly; in Stages 1 and 2, the real trajectories are oversampled six times, giving them about 23\% of the samples.
Epochs are counted over unique trajectories.
Real-robot fine-tuning starts from the pretrained Stage 4 checkpoint, trains Stage 4 only on the 60 demonstrations of the target task without mixing in pretraining data, and keeps AnyTouch2 and the TacRep Encoder frozen as in simulation.

\textbf{Optimization.}
All stages use AdamW with $\beta=(0.9,0.95)$ and weight decay 0.01, cosine decay, and per-group gradient clipping at a norm of 1.0.
Parameters inherited from the base model peak at $5\times10^{-5}$ and the tactile modules at $10^{-4}$.
The loss weights are $\lambda_{\text{vis}}=0.01$ and $\lambda_{\text{tac}}=300$.
Pretraining runs on 8 NVIDIA H100 80GB GPUs and real-robot fine-tuning on 4, both in bfloat16 with two gradient accumulation steps and a micro-batch of 16 per GPU in Stage 1 and 8 per GPU in the other stages, which gives the global batch sizes in Table~\ref{tab:hparams-pretrain}.
No data augmentation is used.

\begin{table}[H]
\centering
\setlength{\belowcaptionskip}{4pt}
\caption{Data and schedule of modest-scale pretraining and real-robot fine-tuning.}
\label{tab:hparams-pretrain}
\small
\begin{tabular}{lccc}
\toprule
 & Data & Length & Global batch \\
\midrule
Stage 1: TacRep Training & OmniViTac + real & 2 epochs & 256 \\
Stage 2: Tactile World Grounding & OmniViTac + real & 2 epochs & 128 \\
Stage 3: Tactile--Action Alignment & OmniViTac & 0.5 epochs & 128 \\
Stage 4: Joint Training & OmniViTac & 3 epochs & 128 \\
Real-robot fine-tuning (Stage 4, per task) & 60 demonstrations & 8,000 steps & 64 \\
\bottomrule
\end{tabular}
\end{table}

\section{Ablation Implementation}
\label{app:ablation-impl}

All variants in Table~\ref{tab:ablation} share the data, architecture, and per-stage hyperparameters of the full model.
The two pathway ablations retrain Stage 4 from the Stage 3 checkpoint of the full model with the attention connections to the removed pathway closed.
The two target-space variants replace the frozen TacRep Encoder with the frozen Cosmos tokenizer or DINOv2 ViT-B, keep the same context and target grids, and retrain Stages 2 to 4.

\section{Baselines}
\label{app:baselines}

We compare with the vision-only and visuo-tactile baselines listed in Table~\ref{tab:main}.
Except for InternVLA-A1, all results are taken from the published UniVTAC evaluations of the respective papers.

\textbf{Vision-only Policies.}
$\pi_{0.5}$~\citep{intelligence2025pi_} is a VLA model built on $\pi_0$ that co-trains on heterogeneous data and decodes action chunks with a flow-matching action expert.
StarVLA-$\alpha$~\citep{ye2026starvla} is a deliberately minimal VLA design that pairs a strong vision-language backbone with a simple action head.
InternVLA-A1~\citep{cai2026internvla} is our base model, a mixture-of-transformers world action model with understanding, visual generation, and action experts (Section~\ref{sec:overview}); we re-evaluate it under the same protocol as TacDyn-WAM so that the gain over the base model is measured under identical conditions.
Xiaomi-Robotics-0~\citep{cai2026xiaomi} is an open-source VLA model trained on large-scale cross-embodiment trajectories and vision-language data with asynchronous real-time execution.
GigaWorld-Policy~\citep{ye2026gigaworld} is an action-centered world action model that learns pixel-action dynamics with optional video generation.
LingBot-VA~\citep{li2026causal} is an autoregressive diffusion framework that learns frame prediction and policy execution in a shared latent space with a mixture-of-transformers backbone.
Fast-WAM~\citep{yuan2026fast} keeps video co-training during training but skips future prediction at test time.
None of these methods uses tactile input.

\textbf{Visuo-tactile Policies.}
ACT+UniVTAC~\citep{chen2026univtac} augments ACT with the UniVTAC tactile encoder, which is pretrained on simulated tactile data with reconstruction and geometric supervision.
VITaL~\citep{george2025vital} pretrains a visuo-tactile encoder and feeds its features to an ACT-style policy.
RDP~\citep{xue2025reactive} is a slow-fast design in which a slow latent diffusion policy predicts action chunks and a fast asymmetric tokenizer performs closed-loop tactile feedback control at high frequency.
TacForcing~\citep{zhou2026tacforcing} replaces the action expert of a VLA with a streaming action expert that conditions actions nearing execution on the tactile observations acquired during execution.
Tactile-WAM~\citep{wu2026tactilewam} is a world action model that predicts future tactile latents with a video VAE and uses asymmetric attention so that video queries do not read tactile keys while action queries do.
The last three methods in this group are additionally pretrained on large-scale tactile data.
FTP-1~\citep{yuan2026ftp} is a generalist tactile policy that maps image, array, and state-based tactile signals from 21 sensors into unified latent tokens through heterogeneous encoders, pretrained on about 3,000 hours of tactile manipulation data.
$N_0$-VTLA~\citep{neoteai2026n0vtla} builds on $\pi_{0.5}$, encodes tactile difference images with a frozen DINOv2, and conditions the action expert on latent tokens that predict the net tactile change over the action chunk.
$N_0$-TWAM~\citep{neoteai2026n0twam} is a 7.2B video diffusion world action model with separate video, tactile, and action experts, which predicts future video and tactile frames in a shared VAE latent space by iterative denoising.
Both $N_0$ models are pretrained on more than 30,000 hours of visuo-tactile robot data from six embodiments and 450 tasks; the 20\% data variant of $N_0$-TWAM is pretrained on one fifth of this corpus.

\section{Real-World Tasks}
\label{app:real-tasks}

\begin{figure}[t]
    \noindent\hspace*{0.0725\linewidth}\includegraphics[width=0.788\linewidth]{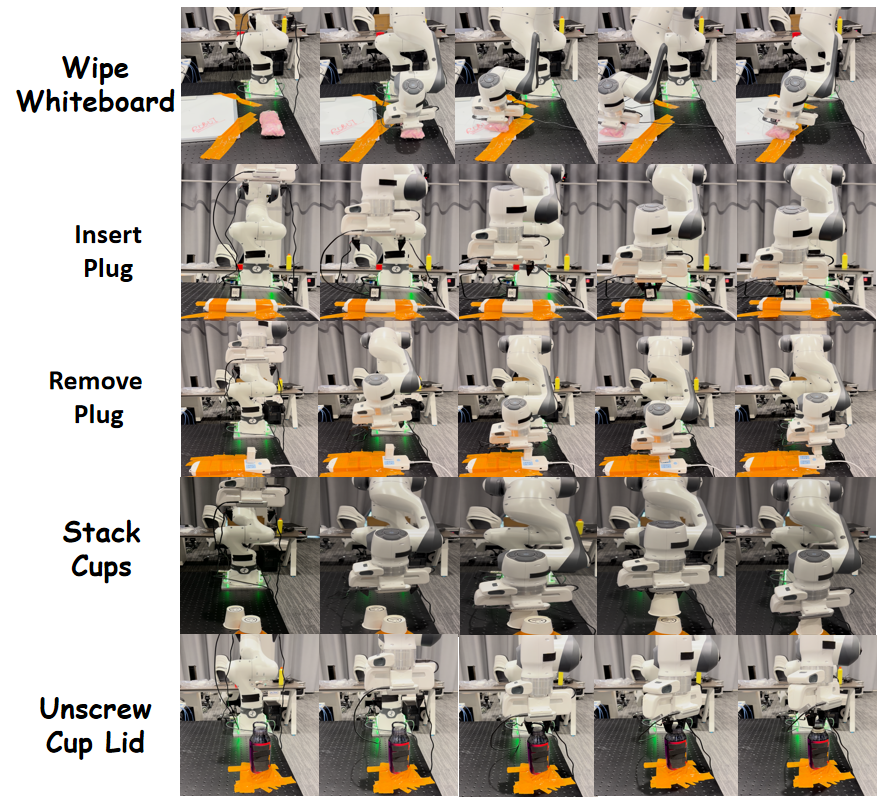}
    \caption{The five real-world tasks. Each row shows one task from left to right over time.}
    \label{fig:real-tasks}
\end{figure}

\textbf{Platform.}
The robot is a Franka Research 3 arm with a parallel gripper.
Each finger carries one Xense vision-based tactile sensor, and the scene is observed by a wrist camera and a third-person camera.
Demonstrations are collected by teleoperation with a leader arm, 60 per task.
Figure~\ref{fig:real-tasks} shows the five tasks.

\begin{figure}[htbp]
    \noindent\hspace*{0.229\linewidth}\includegraphics[width=0.698\linewidth]{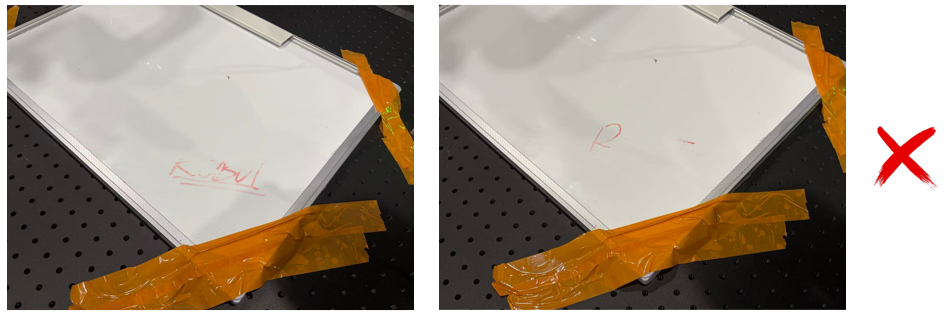}
    \caption{Two failed Wipe Whiteboard trials. Left: the letters are not wiped clean. Right: most of the letters are removed, but part of a letter remains. Both are counted as failures.}
    \label{fig:wipe-fail}
\end{figure}

\textbf{Tasks.}
Each task is specified by a fixed language instruction, and a trial is successful only if the final state meets the criterion below.
\begin{itemize}
\item \emph{Stack Cups} (``pick up the paper cup and stack it on top of the other cup''): grasp a paper cup and nest it onto a second cup. The trial succeeds if the cup rests stably inside the other cup. Misaligned cups tend to jam, and a policy that keeps pushing downward fails, whereas a policy that lifts and re-aligns can recover.
\item \emph{Remove Plug} (``grasp the plug and pull it out of the socket''): grasp a plug seated in a socket and pull it out. The trial succeeds if the plug leaves the socket while still held. The main failure is slip between the fingers and the plug during pulling.
\item \emph{Insert Plug} (``grasp the plug and insert it into the socket''): grasp a plug and insert it into a socket. The trial succeeds if the plug is fully inserted. Approaching at a wrong angle either fails to insert the plug or knocks the socket away.
\item \emph{Unscrew Cup Lid} (``grasp the cup lid and unscrew it''): grasp the lid of a cup and rotate it until it detaches. The trial succeeds if the lid comes off. Rotating without a proper grip can tip the cup over.
\item \emph{Wipe Whiteboard} (``wipe the letters off the whiteboard with the eraser''): grasp an eraser and wipe handwritten letters off a whiteboard. The trial succeeds only if at least 95\% of the letters are removed; letters that are not wiped clean or leave a visible residue, as in Figure~\ref{fig:wipe-fail}, count as a failure.
\end{itemize}

\section{Evaluation Protocol}
\label{app:eval}

\subsection{Simulation Experiments}
\label{app:eval-sim}

\textbf{Tasks.}
UniVTAC~\citep{chen2026univtac} is built on TacEx and Isaac Sim.
The robot is a single-arm Franka with a parallel gripper and a GelSight Mini sensor on each finger.
Table~\ref{tab:tasks} lists the eight tasks.
Expert demonstrations are generated by scripted policies that include randomized failures and contact-based corrections, and success is judged by physics-based criteria that reject excessive gel penetration and slip.

\begin{table}[htbp]
\centering
\setlength{\belowcaptionskip}{4pt}
\caption{The eight UniVTAC tasks.}
\label{tab:tasks}
\small
\begin{tabular}{llp{8.6cm}}
\toprule
Task & Category & Description \\
\midrule
Lift Bottle & Pose reasoning & Grasp a bottle standing against a wall and lift it vertically without hitting the wall. \\
Lift Can & Pose reasoning & Grasp a horizontally placed can of one of three diameters and lift it without slipping. \\
Put Bottle & Pose reasoning & Grasp a standing bottle and place it into the cavity of a shelf. \\
Grasp Classify & Shape perception & Touch two visually similar cylinders with different surface textures, then place each on the target pad of its class. \\
Insert Hole & Contact-rich & Insert a tube into a tilted hole whose orientation must be found by contact. \\
Insert Tube & Contact-rich & Insert a tube into a narrow hole on a tilted surface with a small clearance. \\
Insert HDMI & Contact-rich & Align a held HDMI plug with a random initial rotation and insert it into a fixed socket. \\
Pull-out Key & Contact-rich & Rotate a key in a lock until mechanical resistance is felt, then pull it straight out. \\
\bottomrule
\end{tabular}
\end{table}

\textbf{Rollouts.}
We evaluate one policy per task with 100 rollouts, each starting from a randomized initial configuration drawn by the benchmark.
A rollout is scored with the official UniVTAC success criterion of the task.
By default the full 50-step chunk is executed before the next inference; on Insert Hole and Lift Can we execute the first 32 steps of each chunk, which we found to give more stable contact.
The same protocol is used for our re-evaluation of InternVLA-A1 and for all ablation variants.

\subsection{Real-World Experiments}
\label{app:eval-real}

Each policy is evaluated with 20 trials per task.
Before each trial, the objects are placed at a random position and orientation within the workspace covered by the demonstrations, and success is judged by the criteria in Appendix~\ref{app:real-tasks}.
All policies are trained per task on the same 60 demonstrations.
The three baselines are fine-tuned from their released checkpoints.
TacDyn-WAM without the OmniViTac pretraining stage is trained through the four stages with the simulation settings, including Stage 1 on the tactile clips of the same demonstrations, and TacDyn-WAM (pretrained) is fine-tuned with Stage 4 from the pretrained checkpoint (Appendix~\ref{app:impl-pretrain}).
Both variants predict 50-step action chunks and execute the first 32 steps of each chunk before the next inference.

\textbf{Latency Measurement.}
The latencies in Table~\ref{tab:speed} are measured on the same NVIDIA A100 GPU with batch size 1 and synchronous inference, as the median of 30 runs after warmup, and cover only the GPU computation from the input tensors to the action chunk.
LingBot-VA is run in its two-frame chunk setting, which outputs 40 actions per chunk and is closest to our 50-step chunk, with a history of four chunks.
Its naive setting runs the full video denoising and recomputes the history at every chunk, while its optimized setting follows the deployment configuration of its paper with partial video denoising and a reused key-value cache.
TacDyn-WAM computes the prefix, the memory, and the visual and tactile blocks once per chunk and reuses them across the 10 action denoising steps.

\end{document}